\documentclass[runningheads]{llncs}
\newcommand{\cave}{\texttt{CaVE}}
\newcommand{\cavee}{\texttt{CaVE-E}}
\newcommand{\cavep}{\texttt{CaVE+}}
\newcommand{\caveh}{\texttt{CaVE-H}}
\newcommand{\spo}{\texttt{SPO+}}
\newcommand{\pfyl}{\texttt{PFYL}}
\newcommand{\nce}{\texttt{NCE}}
\newcommand{\spor}{\texttt{SPO+ Rel}}
\newcommand{\pfylr}{\texttt{PFYL Rel}}
\usepackage{graphicx}
\usepackage{subcaption}

\usepackage{bm}
\usepackage[square,sort,comma,numbers]{natbib} 
\usepackage{amsmath}
\usepackage{amssymb}
\usepackage{algorithm}
\usepackage{algorithmic}
\usepackage{color}
\usepackage{hhline}
\usepackage{booktabs}
\DeclareMathOperator*{\argmin}{arg\,min} 

\begin{document}
\title{CaVE: A Cone-Aligned Approach for Fast Predict-then-optimize with Binary Linear Programs}
\titlerunning{CaVE: A Cone-Aligned Approach for Fast Predict-then-optimize}
%
\author{Bo Tang\orcidID{0000-0002-6035-5167} \and
Elias B.~Khalil\orcidID{0000-0001-5844-9642}}
\authorrunning{B. Tang \and E. Khalil}
%
\institute{SCALE AI Research Chair in Data-Driven Algorithms for Modern Supply Chains\\Department of Mechanical and Industrial Engineering, University of Toronto\\
\email{\{botang, khalil\}@mie.utoronto.ca}}
\maketitle              
\begin{abstract}
The \textit{end-to-end predict-then-optimize} framework, also known as \textit{decision-focused learning}, has gained popularity for its ability to integrate optimization into the training procedure of machine learning models that predict the unknown cost (objective function) coefficients of optimization problems from contextual instance information. Naturally, most of the problems of interest in this space can be cast as integer linear programs. In this work, we focus on binary linear programs (BLPs) and propose a new end-to-end training method to predict-then-optimize. Our method, Cone-aligned Vector Estimation (\cave), aligns the predicted cost vectors with the normal cone corresponding to the \textit{true} optimal solution of a training instance. When the predicted cost vector lies inside the cone, the optimal solution to the linear relaxation of the binary problem is optimal. This alignment not only produces decision-aware learning models, but also dramatically reduces training time as it circumvents the need to solve BLPs to compute a loss function with its gradients. Experiments across multiple datasets show that our method exhibits a favorable trade-off between training time and solution quality, particularly with large-scale optimization problems such as vehicle routing, a hard BLP that has yet to benefit from predict-then-optimize methods in the literature due to its difficulty. 

\end{abstract}
%
%
%
\section{Introduction} \label{sec:intro}

Theoretical and experimental results reported over the past few years, starting with~\citet{elmachtoub2021smart} and~\citet{ban2019big}, have demonstrated the need for end-to-end training of Machine Learning (ML) models that predict the cost coefficients of optimization problems. This contrasts with the more traditional two-stage approach, where an ML model is first trained to minimize regression loss for prediction, and then its predictions are applied to new test instances for decision. This conventional approach often leads to substantial \textit{regret} in terms of the quality of the solutions obtained, especially when the training set is small. Given that such predict-then-optimize settings are commonly encountered in many applications (e.g., predicting product demand to manage inventory or travel time on a road network to route trucks), researchers in ML and optimization have proposed a wide range of end-to-end training methods, many of which have been recently surveyed and compared~\cite{tang2022pyepo,sadana2023survey,mandi2023decision}.

With a few exceptions~\cite{elmachtoub2020decision,jeong2022exact}, these methods largely follow the now prevalent mini-batch stochastic gradient descent training algorithm. This process involves the following steps in each iteration: (0) a small batch of training instances is fed into the ML model; (1) the model predicts the cost coefficients; (2) a loss function, incorporating the concept of ``decision error'' such as \textit{regret}, is calculated; (3) the gradients of this loss w.r.t. the parameters of the ML model are computed using backpropagation; and (4) a gradient descent step is employed to update the model parameters. A common feature of many of these methods is the necessity to solve the optimization problem for each training instance at least once in the forward pass (Steps (1, 2)). Since these repeated calls to the (integer) optimization solver represent a significant computational bottleneck, there have been attempts to improve efficiency by replacing them with much cheaper linear optimization calls~\cite{mandi2020smart}, solution caching~\cite{mulamba2020contrastive,mandi2022decision}, and function approximation~\cite{shah2022learning, shah2023leaving, ferber2023surco}. However, these measures often come at a sacrifice in solution quality, as we empirically demonstrate in this paper.

In this work, we are interested in efficient end-to-end training of ML models that predict cost coefficients of challenging binary linear optimization problems, a prime example of which is the Capacitated Vehicle Routing Problem (CVRP)~\cite{toth2014vehicle}. CVRP is defined on a graph comprising customers and a depot, where binary variables represent the edges of the graph and indicate whether or not a vehicle traverses that edge. The number of vehicles, their capacities, and customer demands are fixed and known. Linear constraints capture the requirements of valid tours for all vehicles that start and end at the depot, comply with vehicle capacity limits, and visit each customer exactly once. In the predict-then-optimize setting, each CVRP instance is associated with a ``feature vector'' (e.g., weather conditions, time-of-day, whether it is a holiday or not, etc.) that is predictive of the (unknown) travel times on the graph's edges, which are also the cost coefficients of the objective function, namely the total travel time. CVRP is notoriously hard to solve, even for tens of customers, making end-to-end training of ML models extremely time-consuming.

A distinguishing feature of our \cave{} loss functions is that they do not require solving the original optimization problem during training. Instead, they rely on easier projection problems, continuous and quadratic, which are significantly faster to solve. Our key insight is as follows: By ensuring that the predicted cost vector falls inside a specific cone, namely, one that corresponds to the optimal solution under the true cost vector, we are able to recover this optimal solution. The binding (or active) constraints at the optimal solution define this critical cone. To align the ML model's predicted cost vector with the cone, we need to minimize the angle between the prediction and the cone; this is done through projection onto the cone, our main optimization routine. We show that~\cave{} trains ML models in a fraction of the time required by state-of-the-art methods such as \spo{}~\cite{elmachtoub2021smart} and \pfyl{}~\cite{berthet2020learning} while yielding equally effective cost predictions as measured by regret in unseen test instances.

\section{Related Work} \label{sec:liter}

In the field of operations research, the integration of ML methodologies has emerged as a crucial area of research, significantly reshaping traditional approaches. End-to-end predict-then-optimize, also known as decision-focused learning, effectively utilizes data to tackle optimization problems involving unknown (cost) coefficients.

\subsubsection{KKT-Based Methods.} A notable advancement in this area is the KKT-based method: \citet{amos2017optnet} obtain both optimal solutions and gradients by solving a linear system derived from KKT conditions. \citet{wilder2019melding} adapted the method for linear programs (LPs) by incorporating a small quadratic term into the objective function, while \citet{mandi2020interior} introduced a logarithmic barrier term. Furthermore, \citet{ferber2020mipaal} employ the cutting-plane method, which allows for integer variables. These KKT-based implicit differentiation methods require the use of specialized solving algorithms, which inherently limit their flexibility and often compromise the efficiency of solving problems such as linear programming. Furthermore, aside from the time-intensive cutting-plane method, KKT-based approaches generally struggle to tackle discrete models.

\subsubsection{Black-Box Methods.} In contrast, other methodologies approach the optimization solver as a black box, functioning independently of the solver and algorithm. \citet{elmachtoub2021smart} propose a convex surrogate of \textit{regret} for linear objective functions, which has nonzero subgradients. In contrast to this convex and theoretically sound loss function, \citet{poganvcic2019differentiation} present a linear interpolation, transforming optimization into a piecewise linear function. As a specific case of the interpolation method, \citet{sahoo2022backpropagation} adopt a more direct straight-through estimator by using the negative identity matrix as the surrogate optimization gradient. \citet{niepert2021implicit} extend the interpolated method with perturbations with random noise. In the context of linear objective functions, more perturbation approaches \cite{berthet2020learning, dalle2022learning} involve adding a perturbation to the predicted cost coefficients to smooth the optimization function and further construct a loss function based on duality.

Since the training process involves solving the optimization problem in each iteration, many solutions naturally accumulate as samples. Under the assumption that the feasible region remains fixed, these solutions are all feasible. Therefore, \citet{mulamba2020contrastive} proposes a contrastive loss designed to maximize the distinction between suboptimal solutions in the sample and the optimal solution. Additionally, it utilizes these accumulated solutions as a cache, effectively reducing computational cost. Inspired by the contrastive approach, \citet{mandi2022decision} employed ``learning-to-rank''~\cite{liu2009learning} by ranking the objective value of the cached solutions.

\subsubsection{Function Approximation Methods.} Due to the computational inefficiency often encountered in solving optimization problems, function approximation methods that do not require constraint optimization are appealing. The critical component of function approximation is a learnable surrogate function, which is learned to mimic the original objective or loss function. \citet{shah2022learning, shah2023leaving} samples datasets and employs an additional neural network model. This model is trained to approximate the actual loss of a decision. In doing so, they effectively deploy this approximate loss in end-to-end training. With the approximation method, the complexity of directly dealing with the original function is significantly reduced, allowing for more efficient learning. However, the accuracy of the approximate loss function directly impacts the final model performance. In practice, training a model to learn and approximate a specific function effectively can be a challenging endeavour.

\section{Problem Statement and Preliminaries} \label{sec:prel}

\subsection{Definitions and Notation} \label{subsec:define}

For the sake of clarity, we can define a binary linear program as follows: There are binary decision variables $\bm{w} \in \{\bm{0}, \bm{1} \}^d$. The cost coefficients associated with these decision variables are represented by $\bm{c} \in \mathbb{R}^d$; the constraints are $\bm{A}\bm{w} \leq \bm{b}$, where $\bm{A} \in \mathbb{R}^{k \times d}$ and $\bm{b} \in \mathbb{R}^k$:

\begin{equation}
  \begin{aligned}
    \min_{\bm{w}} \quad & \bm{c}^\intercal \bm{w} \\
    \textrm{s.t.} \quad & \bm{A}\bm{w} \leq \bm{b}, \\
    & \bm{w} \in \{\bm{0}, \bm{1} \}^d. 
    \label{eq:blp}
  \end{aligned}
\end{equation}

Let $\bm{\Omega}$ represent the feasible region of the problem; then the optimal solution is expressed as $\bm{w}^*(\bm{c}) = \argmin_{\bm{w} \in \bm{\Omega}} \bm{c}^\intercal \bm{w}$. Given its computational complexity, the process of determining $\bm{w}^*(\bm{c})$ can be extremely time-consuming, especially when the number of decision variables and constraints is large.

In the predict-then-optimize setting, the coefficients $\bm{c}$ are unknown and linked to a feature vector $\bm{x} \in \mathbb{R}^p$. This relationship facilitates the use of an ML model $g(\bm{x}, \bm{\theta})$ to estimate the predicted coefficients $\hat{\bm{c}}$. In this model, $\bm{\theta}$ represents the learnable parameters, which are adjusted to minimize a decision loss $\mathcal{L}(\cdot)$, a metric that quantifies the discrepancy between the true optimal solution $\bm{w}^*(\bm{c})$ and the solution derived from the prediction $\hat{\bm{c}}$.

\subsection{Metric} \label{subsec:metric}

To measure the quality of the decision, \textit{ regret} is introduced by \citet{elmachtoub2021smart}, which is defined as the absolute gap between the objective value of the solution $\bm{w}^*(\hat{\bm{c}})$ obtained using the predicted coefficients $\hat{\bm{c}}$, and the optimal value of the solution $\bm{w}^*(\bm{c})$ obtained using the actual coefficients $\bm{c}$. It is expressed in the following equation:
\begin{equation}
  \begin{aligned}
    \mathcal{L}_{\text{Regret}}(\hat{\bm{c}}, \bm{c}) = \bm{c}^\intercal \big( \bm{w}^* (\hat{\bm{c}}) - \bm{w}^* (\bm{c})\big).
  \end{aligned}
\end{equation}

In line with \citet{elmachtoub2021smart}, we adopt \textit{normalized regret}, which serves as an adjusted metric that takes into account the scale of the problem, providing a more standardized and comparable measure with $n$ samples:
\begin{equation}
  \begin{aligned}
    \frac{\sum_{i=1}^{n}{\mathcal{L}_{\text{Regret}} \left( \hat{\bm{c}}_i, \bm{c}_i \right) }}{\sum_{i=1}^n { \left| \bm{c}_i^\intercal \bm{w}_i^*(\bm{c}_i) \right| }}.
  \end{aligned}
\end{equation}
The regret is best understood as follows: If a method records a test regret of $0.07$ for example, it means that it produces solutions that are $7\%$ worse than the true optimal solutions under the true but unknown cost vectors.

\section{Methodology} \label{sec:method}

\subsection{Optimal Cones and Subcones} \label{subsec:cone}

\begin{figure}[htbp!]
    \centering
    \includegraphics[width=0.4\textwidth]{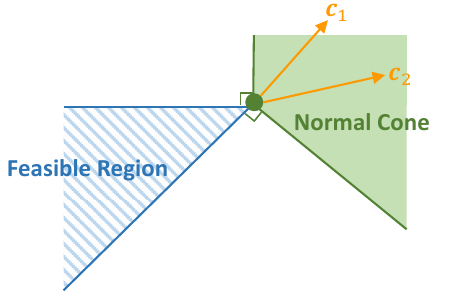}    
    \caption{Illustration of a normal cone: the cost vectors $\bm{c}_1$ and $\bm{c}_2$ produce the same optimal solution if and only if they lie within this cone.}
    \label{fig:cone}
\end{figure}

\begin{figure}[htbp]
    \centering
    \includegraphics[width=\textwidth]{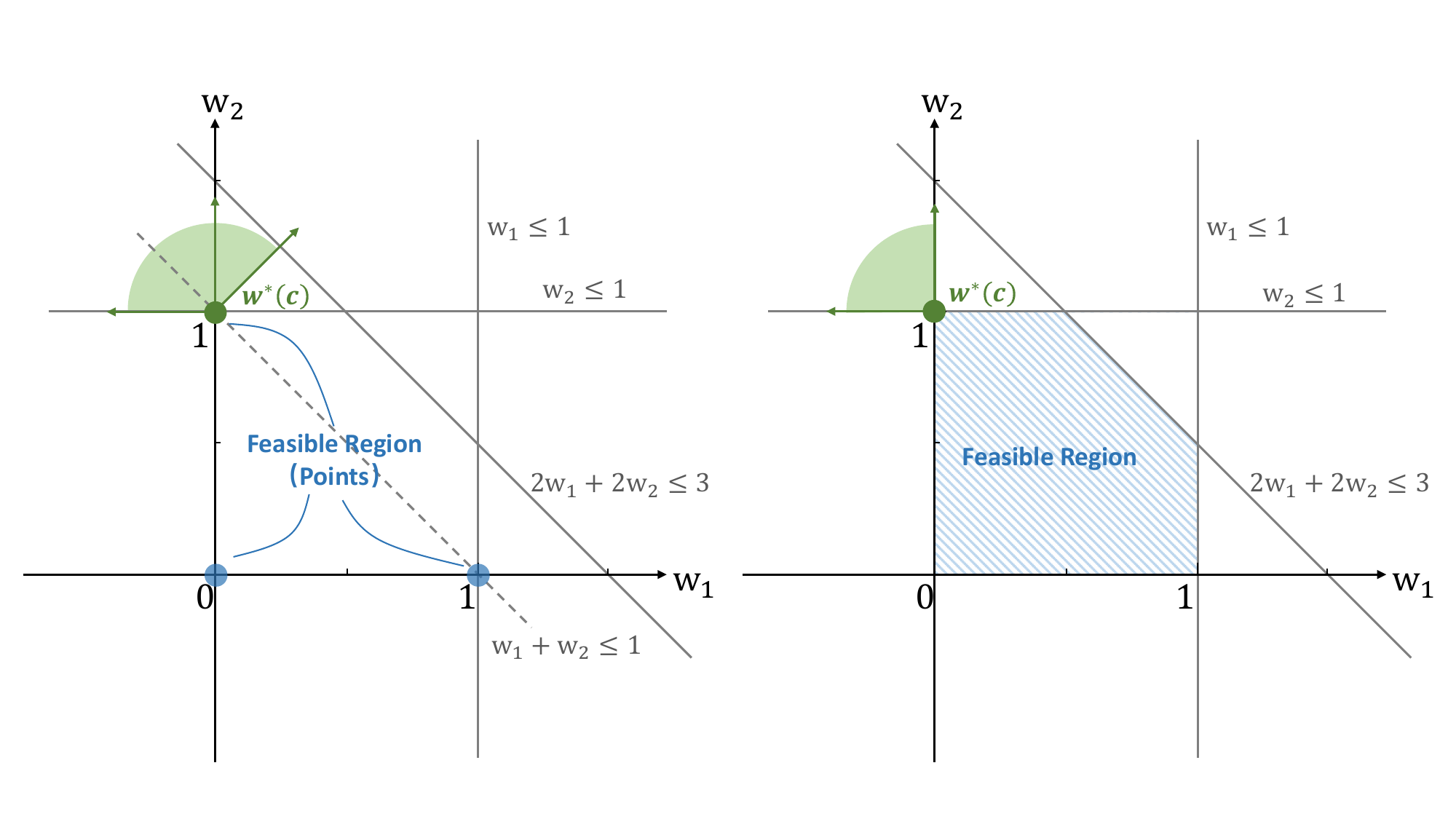}    
    \caption{Illustration of the optimal cone and optimal subcone: On the left, the green cone is the optimal cone of a BLP. On the right, the green cone is a subset of the left cone and the optimal cone of the LP relaxation of the BLP on the left.}
    \label{fig:subcone}
\end{figure}

In (continuous) LP, one can associate a normal cone as \textit{optimal cone} with a given cost vector $\bm{c}\in\mathbb{R}^d$. Within this cone, all cost coefficients yield the same optimal solution $\bm{w}^*(\bm{c})$. As depicted on the right of Fig.~\ref{fig:cone}, the construction of the normal cone leverages the conical combination of the binding constraints $\widetilde{\bm{A}}(\bm{c})$ at $\bm{w}^*(\bm{c})$: each vector within the cone can be represented as a non-negative combination of the binding constraints coefficients $\bm{a}_j \in \widetilde{\bm{A}}(\bm{c})$, which can be written as 
$$\sum_{\bm{a}_j \in \widetilde{\bm{A}}(\bm{c})} {\lambda}_i \bm{a}_j, \forall \bm{\lambda} \geq \bm{0},$$
Now consider the case of a binary linear program. If linear cuts $\bm{A}^{\prime}(\bm{c})$ correspond to the convex hull of integer points of the BLP, the same logic to obtain the optimal cone of an integer solution would have been applied. This parallel is illustrated on the right side of Figure~\ref{fig:subcone} for a BLP optimal cone, defined as
$$\mathcal{C}^*(\bm{c}) = \left\{ \bm{v} \in \mathbb{R}^d : \bm{v} = \sum_{\bm{a}_j \in \bm{A}^{\prime}(\bm{c})} \lambda_i \bm{a}_i, \forall \bm{\lambda} \geq \bm{0} \right\}.$$
Here, the cone is delineated by the cuts $w_1\geq 0$, $w_2 \leq 1$ and $w_1+w_2\leq 1$. However, we typically do not operate on the convex hull of a BLP, necessitating an alternate definition for the cones with which we will attempt to align the predictions.

To this end, we keep the normal cone of a BLP as the \textit{optimal subcone}:
$$\mathcal{SC}^*(\bm{c}) = \left\{ \bm{v} \in \mathbb{R}^d : \bm{v} = \sum_{\bm{a}_j \in \widetilde{\bm{A}}(\bm{c})} \lambda_i \bm{a}_i, \forall \bm{\lambda} \geq \bm{0} \right\}.$$
For a BLP with an optimal solution $\bm{w}^*(\bm{c})$, the optimal subcone is the optimal cone of the same solution, but with the LP relaxation of the BLP instead. In Fig.~\ref{fig:subcone}, the cone on the left side is the optimal cone for the BLP; the cone on the right side is the optimal subcone. Since $\mathcal{SC}^*(\bm{c}) \subset \mathcal{C}^*(\bm{c})$, all rays within the subcone also belong to the optimal cone. Although the LP relaxation leads to an expanded feasible region, all binary feasible solutions of the original BLP remain feasible vertices in the LP. As such, recovering an optimal subcone for a given solution $\bm{w}^*(\bm{c})$ is trivial:
for a BLP problem defined as in Equation~\ref{eq:blp}, its optimal subcone is the set of vectors that can be expressed as nonnegative combinations of the coefficient vectors $\bm{a}_j \in \widetilde{\bm{A}}(\bm{c})$ that satisfy $\bm{a}_j\bm{w}^*(\bm{c}) = b_j$, including the variable bounds for which we have either $w_i^*(\bm{c}) = 0$ or $w_i^*(\bm{c})=1$. 

\subsection{Cone-aligned Vector Estimation: three variants} \label{subsec:cave}

Cone-aligned Vector Estimation (\cave{}) is an approach specially designed for end-to-end training in BLP. The core idea is to leverage the optimal subcone defined in Sec.~\ref{subsec:cone}. This approach aims to train an ML model so that the predicted cost coefficients reside within this optimal subcone.

To drive the predicted vector $\hat{\bm{c}}$ into the subcone, the ML model is trained to reduce the \textit{angle} $\phi$ between the cost prediction and the subcone. Accordingly, the loss $\mathcal{L}_{\text{CaVE}} (\cdot)$ to be minimized can be defined as the negative cosine similarity between the prediction and its projection onto the subcone, namely:
\begin{equation}
    \mathcal{L}_{\text{CaVE}} (\hat{\bm{c}}, \widetilde{\bm{A}}(\bm{c})) 
    = -\text{cosine\_similarity}(\hat{\bm{c}}, \bm{p}_{\hat{\bm{c}}})
    = - \dfrac{\hat{\bm{c}}^\intercal \bm{p}_{\hat{\bm{c}}}}{\lVert\hat{\bm{c}}\rVert \lVert\bm{p}_{\hat{\bm{c}}}\rVert},
    \label{eq:loss}
\end{equation}
where, assuming there are $m$ binding constraints, i.e., $\widetilde{\bm{A}}(\bm{c}) \in \mathbb{R}^{m\times d}$, the projection writes:
\begin{equation}
    \bm{p}_{\hat{\bm{c}}} = {\widetilde{\bm{A}}(\bm{c})}^\intercal \bm{\lambda}^{*}, \quad
    \bm{\lambda}^{*} = \argmin_{\bm{\lambda}\geq\bm{0}} \lVert {\widetilde{\bm{A}}(\bm{c})}^\intercal \bm{\lambda} - \hat{\bm{c}}\rVert^2.
    \label{eq:proj}
\end{equation}

\begin{algorithm}
\caption{Cone-aligned Vector Estimation (\cave{})}
\label{alg:cave}
\begin{algorithmic}[1]
\REQUIRE Pairs of feature vectors and binding constraints $\{(\bm{x}^i,  \widetilde{\bm{A}}^i)\}_{i=1}^{n}$ for $n$ training instances; learning rate $\alpha > 0$
\STATE Initialize model parameters $\bm{\theta}$
\FOR{each training epoch}
    \FOR{each batch of training samples $(\bm{x}, \widetilde{\bm{A}})$}
        \STATE Predict cost coefficient $\hat{\bm{c}} \gets g(\bm{x}, \bm{\theta})$
        \STATE Compute projection $\bm{p}_{\hat{\bm{c}}}$ with  quadratic program~\eqref{eq:proj}
        \STATE Compute cosine similarity loss $\mathcal{L}_{\text{CaVE}} (\hat{\bm{c}}, \widetilde{\bm{A}})$~\eqref{eq:loss}
        \STATE Compute the gradient $\nabla_{\theta}\mathcal{L}_{\text{CaVE}} (\hat{\bm{c}}, \widetilde{\bm{A}})$ with backpropagation
        \STATE Update ML model parameters $\bm{\theta} \gets \bm{\theta} - \alpha \nabla_{\theta}\mathcal{L}_{\text{CaVE}} (\hat{\bm{c}}, \widetilde{\bm{A}})$
    \ENDFOR
\ENDFOR
\RETURN $g(\cdot, \bm{\theta})$
\end{algorithmic}
\end{algorithm}

When the alignment is precise, i.e., the predicted cost vector falls within the correct optimal subcone, the \cave{} loss achieves its minimum value of $-1$, indicating an optimal decision. Although our method still requires a quadratic program (QP) to compute the projection of the prediction values during each training iteration, it effectively circumvents the need to solve the more challenging binary linear program. Algorithm~\ref{alg:cave} presents a detailed, step-by-step description of the \cave{} training process.

\begin{figure}[htbp]
    \centering
    \includegraphics[width=1.0\textwidth]{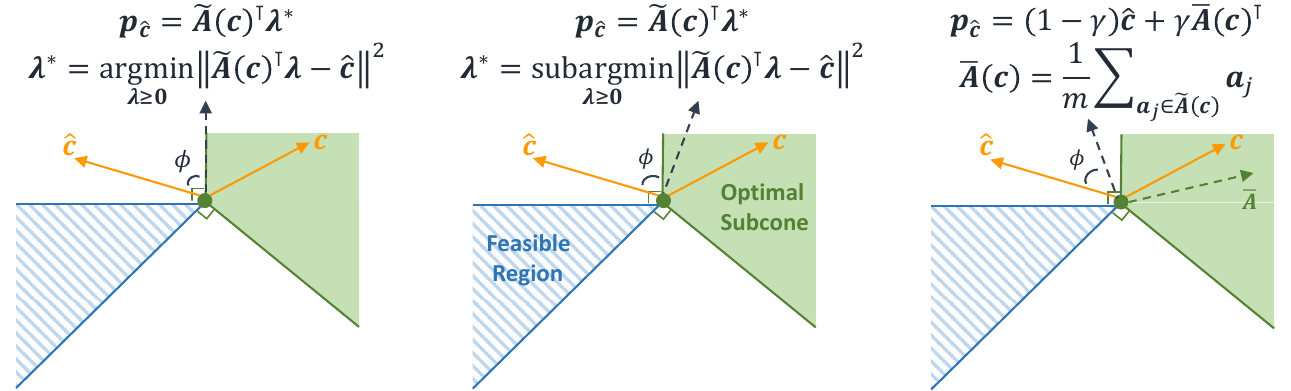}    
    \caption{Illustration of the three projections: Exact projection on the left, inner projection in the middle, and heuristic projection on the right.}
    \label{fig:proj}
\end{figure}

Figure~\ref{fig:proj} illustrates the three types of projection that can be employed in the \cave{} framework. The \textit{exact projection} projects the cost coefficients $\hat{\bm{c}}$ directly onto the surface of the optimal subcone; this is the approach that we have just laid out in Algorithm~\ref{alg:cave}. The \textit{inner projection} ensures that the projected vector lies strictly within the subcone. The \textit{heuristic projection} is an approximation of the true projection onto the optimal subcone, used to reduce the computational cost. 
We will detail these three variants next. 

\subsubsection{\cavee{} with exact projection.} \cavee{} performs exact projection, wherein the optimal solution of the quadratic programming problem~\eqref{eq:proj} is computed to locate the projection on the surface of the cone, as illustrated on the left in Fig.~\ref{fig:proj}. Nevertheless, the \cavee{} method encounters a significant drawback due to its projection onto the face of the cone and the use of cosine similarity as the loss function. This approach results in the vanishing of gradients as the predicted cost vector nears the surface of the optimal subcone but is yet to enter it. Experimental evidence in Section~\ref{sec:exper} corroborates this issue -- the regret of \cavee{} is typically higher than existing end-to-end methods, necessitating a modification.

\subsubsection{\cavep{} with inner projection.} Due to the issue of vanishing gradients associated with \cavee{}, \cavep{} replaces the exact projection with what we refer to as ``inner projection''. The goal is to obtain a projection of the predicted cost vector that lies \textit{inside the subcone}. As all optimal projections will lie on a face of the subcone, we thus require a suboptimal solution to the projection problem. This is readily achieved by simply limiting the number of iterations in the quadratic programming solver and thus terminating prematurely. Since the solver uses the primal-dual interior point method, the feasibility is guaranteed at each iteration.  In our experiments, the maximum number of iterations of the QP solver used by \cavep{} was set to $3$. A suboptimal projection will lie inside the optimal subcone, resulting in nonzero loss and a strong gradient signal that will push the ML model parameters to produce predictions that move toward the inside of the subcone. Compared to exact projection, this approach is more computationally efficient with fewer iterations of the interior point method. The inner projection is illustrated in the middle of Fig.~\ref{fig:proj}.

\subsubsection{\caveh{} with a mix of inner and heuristic projections.} To alleviate the computational burden of repeated QP solving in both~\cavee{} and \cavep{}, a hybrid strategy is employed in \caveh{}. We interleave inner projections (obtained with a QP just as in~\cavep{}) with much cheaper heuristic projections. Unlike exact and inner projections, the heuristic projection does not necessitate solving a quadratic program and instead requires a simple convex combination of $\overline{\bm{A}}(\bm{c})$ and $\hat{\bm{c}}$ with weight $\gamma \in [0,1]$:
$$\bm{p}_{\hat{\bm{c}}}=(1-\gamma)\hat{\bm{c}} + \gamma{\overline{\bm{A}}(\bm{c})}^\intercal,$$
where $\overline{\bm{A}}(\bm{c})=\frac{1}{m}\sum_{\bm{a}_j \in \widetilde{\bm{A}}(\bm{c})}{\bm{a}_j}$ is the average of all binding constraints. As illustrated on the right in Fig.~\ref{fig:proj}, it is crucial to note that the heuristic projection is not guaranteed to be in the optimal subcone, but it still ensures that the cost coefficient vector is pushed in the direction of the optimal subcone. 
With probability $\beta < 0.5$, i.e., in the minority of the iterations of Algorithm~\ref{alg:cave},~\caveh{} performs an inner projection via QP. With probability $(1-\beta)$, the heuristic projection is used instead, without any optimization required. In our experiments, $\gamma$ and $\beta$ are set to $0.2$ and $0.3$, respectively, and were not tuned any further for performance.

\begin{table}[htpb!]
\centering
\begin{tabular}{l@{\hskip .2in}c}
\toprule
Method & Iteration cost\\ \midrule
\pfyl & $K\times$BLP\\
\spo & $1\times$BLP\\
\nce & $\beta\times$BLP\\ 
\cavee & $1\times$QP\\
\cavep & $1\times$QP (partial)\\
\caveh & $\beta\times$QP (partial)\\ 
\bottomrule\\
\end{tabular}
\caption{Comparison of state-of-the-art predict-then-optimize methods \spo{}, \pfyl{}, and \nce{} with the three~\cave{} variants w.r.t. ``Iteration cost'' (per training instance), i.e., the frequency of time-consuming solver calls required to compute the loss of a method during gradient descent. The methods are sorted in decreasing order of their iteration costs. \spo{} requires solving the BLP with predicted costs and \pfyl{} requires solving the $K$ (typically 1-5) BLPs, whereas \nce{} with solution cache needs to solve only a small fraction ($100 \times \beta\%$) of the BLPs. \cave{} methods require solving a single QP, or partially solving a single QP in ($100 \times \beta\%$) of the iterations.}
\label{tab:comparison}
\end{table}


\subsubsection{Comparison with existing methods.}
Table~\ref{tab:comparison} summarizes the computational cost of training predict-then-optimize models using \spo{}~\cite{elmachtoub2021smart}, \pfyl{}~\cite{berthet2020learning}, \nce{} \cite{mulamba2020contrastive} and the three~\cave{} variants. We choose \spo{} and \pfyl{} as they represent the state-of-the-art based on an extensive evaluation carried out in~\cite{tang2022pyepo}, though our experiments will include \nce{}  with solution cache, a fast method, as well as accelerated variants of \spo{} and \pfyl{}. The table shows that~\caveh{} and~\nce{} are the fastest to train, whereas \pfyl{} and \spo{} exhibit similar costs, particularly when \pfyl{}'s number of random perturbations of the predicted cost vector is set to $K=1$.~\cavep{} sits between~\cavee{} and~\caveh{} in terms of training cost. We will empirically examine these theoretical complexities in the next section.

\section{Benchmark Datasets} \label{sec:data}

We utilized synthetic datasets~\cite{tang2022pyepo} for our experiments. The synthetic dataset, denoted as $\mathcal{D}$, comprises feature vectors $\bm{x}$ and the corresponding cost coefficients $\bm{c}$. Each feature vector $\bm{x}$ adheres to a standard Gaussian distribution, and the associated cost vector $\bm{c}$ is derived from a polynomial function of $\bm{x}$, with added random noise.

\subsubsection{Shortest Path.} The shortest path problem is a fundamental task in graph theory and network analysis, where the objective is to find a minimum weight path from a fixed source node to a fixed target node. Our instances are based on $5 \times 5$ grid networks (SP5) where the source is the node in the northwest corner of the grid and the target is the node in the southeast corner. The cost coefficient $c_{ij}$ comes from 
$$\Bigg[\frac{1}{{3.5}^\text{deg}} \Bigg(\frac{1}{\sqrt{5}}(\mathcal{B} \bm{x}_i)_j + 3\Bigg)^\text{deg} + 1\Bigg] \cdot \epsilon_{ij},$$
where feature size is $5$, $\mathcal{B}$ follows a Bernoulli distribution, $\epsilon_{ij}$ is random noise uniformly distributed between 0.5 and 1.5, and $\text{deg}$ is the polynomial degree of feature mapping. This is a standard and easy task introduced by~\citet{elmachtoub2021smart}.

\subsubsection{Traveling Salesperson and Vehicle Routing.} We utilized the traveling salesperson problem (TSP) dataset. TSP, known for its NP-hardness, is a classic problem in combinatorial optimization where the goal is to find the shortest tour that visits a set of locations exactly once and returns to the original starting point. In our study, we do not only test TSP instances with 20 and 50 nodes, but also explore the same graphs under the much more challenging CVRP problem. In this dataset, the cost coefficient $c_{ij}$ comes from two parts: The first part consists of the Euclidean distances among the nodes, and the second part resembles the shortest path problem, formulated as $$\Bigg( \frac{1}{\sqrt{10}}( \mathcal{B} \bm{x}_i)_j + 3 \Bigg)^\text{deg} \cdot \epsilon_{ij},$$ where the feature vector $\bm{x} \in \mathbb{R}^{10}$. For CVRP, the capacity of each vehicle is set at 30, and each customer's demand follows a uniform distribution from 0 to 10. To our knowledge, our work is the first to use a CVRP in a predict-then-optimize setting. As will become apparent in the next section, this is due to the time required to solve the BLP representing the CVRP, making \spo{} and \pfyl{} extremely time-consuming as per Table~\ref{tab:comparison}.

\section{Experimental Results} \label{sec:exper}

\subsection{Experimental setup}

The experiments were designed to evaluate both the training time and normalized regret, with the size of both the training and test sets being 1,000 instances. To account for randomness in data generation and stochastic model training, $5$ or $10$ random seeds are used to generate and train training/test sets and the corresponding ML models. Our comparative analysis encompassed several methods, including three variants of \cave{}, a two-stage approach (least-squares regression on cost vectors), \spo{}, \pfyl{}, and \nce{}. A linear model was used as the cost prediction model class for all the aforementioned approaches; this is a standard choice in the literature starting with~\cite{elmachtoub2021smart}.  In the case of \pfyl{}, the size of random sampling is fixed at $K=1$;  for \nce{}, the solving ratio is set at $\beta = 5\%$.

\begin{figure}[htbp]
    \centering
    \begin{subfigure}{0.45\textwidth}
        \centering
        \includegraphics[width=\linewidth]{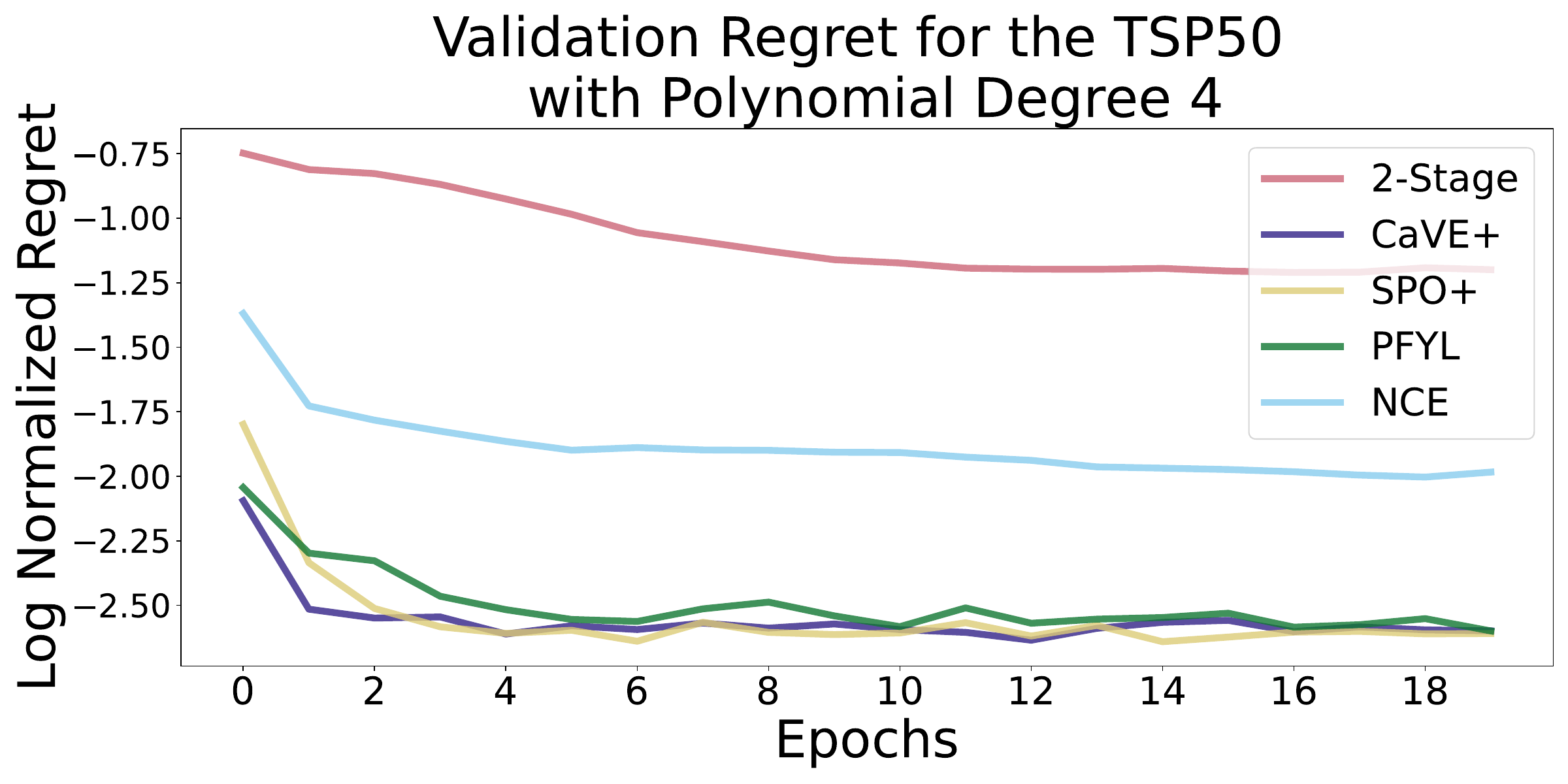}
    \end{subfigure}
    \begin{subfigure}{0.45\textwidth}
        \centering
        \includegraphics[width=\linewidth]{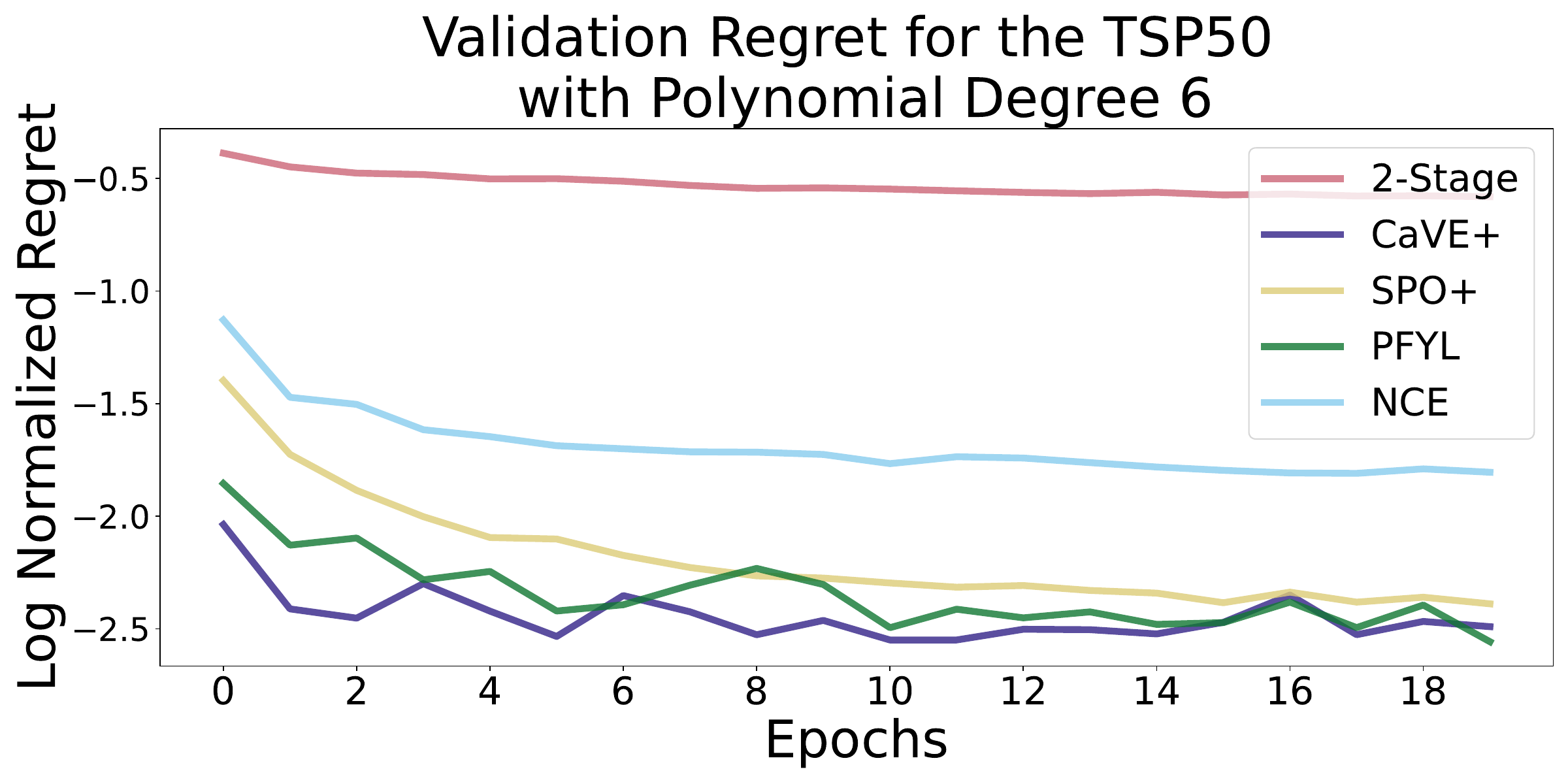}
    \end{subfigure}
    \caption{Validation regeret curves for TSP50 with a polynomial of degree 4 (left) and 6 (right). The vertical axis represents the average normalized regret values of each of the five methods for the validation dataset with $1000$ instances.}
    \label{fig:loss}
\end{figure}

The Adam optimizer was used for gradient descent for all our experiments. For SP5, a learning rate of $0.01$ was used with all methods for 10 epochs, except for the two-stage and \nce{} method which allotted 20 epochs at the same learning rate. For both TSP and CVRP, the only modification was to increase the learning rate to $0.05$, maintaining the number of epochs at 10. In selecting the number of epochs, careful consideration was given to the convergence of each model, ensuring that our results are not skewed by over- or under-training. The loss curves in Fig.~\ref{fig:loss} show that all methods exhibit a convergent behavior, with~\cave{} methods doing so earlier than the baselines. Note that the 2-Stage models are trained essentially to its optimality as they are convex optimization problems.

Our numerical experiments were conducted using Python v3.9.6 on a system with 8 Intel E5-2683 v4 CPUs and 32GB memory. We utilized SciPy~\cite{virtanen2020scipy} v1.11.2 and Clarabel 0.6.0 for QP solving, Gurobi~\cite{gurobi} 10.0.3 for BLP, and PyTorch~\cite{paszke2019pytorch} v2.0.1 with PyEPO~\cite{tang2022pyepo} v0.3.6 for end-to-end training, where PyEPO provided implementations for \spo{}, \pfyl{} and \nce{}. Our code is available at~\url{https://github.com/khalil-research/CaVE}.

\subsection{Results}
Tables~\ref{tab:sp},~\ref{tab:tsp20},~\ref{tab:tsp50},~\ref{tab:vrp20}, and~\ref{tab:vrp30} summarize the results. The tables correspond to the Shortest Path problem on $5 \times 5$ grid (SP5), TSP with 20 nodes (TSP20), TSP with 50 nodes (TSP50), CVRP with 20 customers (CVRP20), and CVRP with 30 customers (CVRP30), respectively. Each table contains two sub-tables: (a) reports the normalized regret metric defined in Section~\ref{subsec:metric} on 1,000 test instances and (b) reports the training time in seconds; both quantities are averages over $5$ random seeds for CVRP20 and $10$ random seeds for others with standard deviation, while the experiments for CVRP30 is not repeated due to running time. Each sub-table has two rows, one corresponding to datasets that use a degree-4 polynomial in the (unknown) mapping from instance features to cost coefficients, and the other for a degree-6 polynomial; we refer to Section~\ref{sec:data} for details on the role of the polynomial in the function that we are attempting to learn, but note that the higher the degree the more difficult the learning task. In each sub-table, we bold the best-performing method, excluding the 2-Stage method as it is often fast to train, but has much worse regret than end-to-end methods.

\subsubsection{Shortest Path.} This is the easiest BLP (in fact, LP) we will look at. It serves as a sanity check for any new method in this space. Table~\ref{tab:sp} shows that~\spo{} and~\pfyl{} achieve the lowest test regrets, closely followed by~\caveh{} and~\cavep{}. The latter two trains roughly $10$ times faster than~\spo{} and~\pfyl{}, already substantiating our claim that QP solving is faster in solving the optimization problem itself, even for small-scale polynomial-time solvable shortest path problems. In addition, although \nce{} is fast, it has a higher regret than others.

\begin{table}[ht]
  \centering
  \caption{Experimental Results for SP5}
    \begin{subtable}{\textwidth}
      \centering
      \caption{Average Test Normalized Regret (\%) with Standard Deviation}
      \setlength{\tabcolsep}{5pt}
      \resizebox{\columnwidth}{!}{
      \begin{tabular}{r|r|r|r|r|r|r|r}
       \toprule
        \textbf{Methods}   & 2-Stage  & \cavee{} & \cavep{} & \caveh{} & \spo{}          & \pfyl{}         &\nce{}  \\ \midrule
        \textbf{Deg $4$}   & $8.82 \pm 1.15$  & $10.73 \pm 1.54$    & $8.39 \pm 0.95$   & $8.35 \pm 0.88$  & $7.79 \pm 1.00$  & $\bm{7.68} \pm 0.99$ & $11.34 \pm 1.11$ \\
        \textbf{Deg $6$}   & $12.58 \pm 2.14$ & $11.30 \pm 1.30$ & $8.89 \pm 0.90$ & $8.84 \pm 1.00$ & $\bm{7.72} \pm 1.11$ & $7.86 \pm 0.96$ & $13.78 \pm 1.58$  \\
      \bottomrule
      \end{tabular}
}
    \end{subtable} 

    \begin{subtable}{\textwidth}
      \centering
      \caption{Average Training Time (Sec) with Standard Deviation}
      \setlength{\tabcolsep}{7pt}
\resizebox{\columnwidth}{!}{
      \begin{tabular}{r|r|r|r|r|r|r|r}
      \toprule
        \textbf{Methods}   & 2-Stage  & \cavee{}  & \cavep{} & \caveh{}       & \spo{} & \pfyl{}  &\nce{} \\ \midrule
        \textbf{Deg $4$}   & $1.52 \pm 0.14$ & $4.64 \pm 0.09$     & $4.89 \pm 0.12$ & $\bm{2.57} \pm 0.19$  & $17.64 \pm 0.12$ & $18.52 \pm 0.31$ & $4.50 \pm 0.48$\\
        \textbf{Deg $6$}   & $1.38 \pm 0.13$ & $3.52 \pm 0.11$ & $3.72 \pm 0.14$ & $\bm{2.39} \pm 0.19$ & $18.68 \pm 0.40$ & $17.78 \pm 0.13$ & $4.38 \pm 0.42$\\
    \bottomrule
      \end{tabular}
}
\end{subtable}
\label{tab:sp}
\end{table}

\subsubsection{TSP.}
For the TSP, \spo{}, \pfyl{} and~\nce{} employ the Dantzig–Fulkerson–Johnson (DFJ) formulation~\cite{dantzig1954solution} to solve this BLP efficiently. For both TSP20 and TSP50, \cavep{} achieves the best time-regret trade-off across all methods: its regret is the best or second-best across all methods and its training time is the second-best after~\caveh{}. As shown in part (b) of Tables~\ref{tab:tsp20} and~\ref{tab:tsp50},~\cavep{} trains in roughly half the time of~\spo{} and~\pfyl{}, achieving the same or even better test regret. Additionally, we also compared our methods with \spor{} and \pfylr{}, which employ a linear relaxation of the BLP during training, as detailed in the Appendix~\ref{subapdx:tsprel}. While they do reduce the training time of vanilla~\spo{}/~\pfyl{} by a bit, this typically comes at an increase in regret. Similarly,~\nce{} has a prohibitively high regret.

\begin{table}[ht]
  \centering
  \caption{Experimental Results for TSP$20$}
    \begin{subtable}{\textwidth}
      \centering
      \caption{Average Test Normalized Regret (\%) with Standard Deviation}
      \setlength{\tabcolsep}{5pt}
       \resizebox{\columnwidth}{!}{
      \begin{tabular}{r|r|r|r|r|r|r|r}
      \toprule
        \textbf{Methods}     & 2-Stage  & \cavee{} & \cavep{} & \caveh{} & \spo{}   & \pfyl{} &\nce{} \\ \midrule
        \textbf{Deg $4$}     & $12.12 \pm 0.89$  & $7.35 \pm 0.40$  & $6.20 \pm 0.24$  & $7.69 \pm 0.33$  & $\bm{5.95} \pm 0.16$ & $6.56 \pm 0.21$  & $12.21 \pm 0.88$ \\
        \textbf{Deg $6$}     & $21.32 \pm 1.81$  & $8.01 \pm 0.45$  & $\bm{6.97} \pm 0.37$  & $9.52 \pm 0.64$  & $7.48 \pm 0.36$  & $7.41 \pm 0.37$  & $14.31 \pm 0.40$ \\
      \bottomrule
      \end{tabular}}
    \end{subtable} 

    \begin{subtable}{\textwidth}
      \centering
      \caption{Average Training Time (Sec) with Standard Deviation}
      \setlength{\tabcolsep}{7pt}
      \resizebox{\columnwidth}{!}{
      \begin{tabular}{r|r|r|r|r|r|r|r}
      \toprule
        \textbf{Methods}   & 2-Stage  & \cavee{} & \cavep{} & \caveh{} & \spo{}   & \pfyl{} &\nce{} \\ \midrule
        \textbf{Deg $4$}   & $1.52 \pm 0.10$  & $113.56 \pm 3.16$ & $107.15 \pm 3.80$ & $27.06 \pm 2.17$ & $175.23 \pm 4.95$ & $220.21 \pm 24.20$ & $\bm{25.92} \pm 4.23$ \\
        \textbf{Deg $6$}   & $1.53 \pm 0.19$  & $158.66 \pm 9.65$ & $102.19 \pm 10.38$ & $30.17 \pm 2.62$ & $185.13 \pm 7.44$ & $185.02 \pm 5.09$ & $\bm{25.48} \pm 3.66$ \\
      \bottomrule
      \end{tabular}}
    \end{subtable}
    \label{tab:tsp20}
\end{table}

\begin{table}[ht]
  \centering
  \caption{Experimental Results for TSP$50$}
    \begin{subtable}{\textwidth}
      \centering
      \caption{Average Test Normalized Regret (\%) with Standard Deviation}
      \setlength{\tabcolsep}{5pt}
\resizebox{\columnwidth}{!}{
      \begin{tabular}{r|r|r|r|r|r|r|r}
      \toprule
        \textbf{Methods}     & 2-Stage  & \cavee{} & \cavep{} & \caveh{} & \spo{}   & \pfyl{} &\nce{} \\ \midrule
        \textbf{Deg $4$}     & $28.16 \pm 1.08$ & $15.19 \pm 0.65$ & $7.69 \pm 0.22$ & $9.59 \pm 0.44$ & $\bm{7.57} \pm 0.20$ & $8.03 \pm 0.23$ & $14.31 \pm 0.40$ \\
        \textbf{Deg $6$}     & $52.61 \pm 2.36$ & $23.25 \pm 2.41$ & $\bm{8.57} \pm 0.38$ & $11.28 \pm 0.80$ & $10.26 \pm 0.46$ & $9.00 \pm 0.52$ & $17.12 \pm 0.48$ \\
      \bottomrule
      \end{tabular}
      }
    \end{subtable} 

    \begin{subtable}{\textwidth}
      \centering
      \caption{Average Training Time (Sec) with Standard Deviation}
      \setlength{\tabcolsep}{7pt}
\resizebox{\columnwidth}{!}{
      \begin{tabular}{r|r|r|r|r|r|r|r}
      \toprule
        \textbf{Methods}   & 2-Stage  & \cavee{} & \cavep{} & \caveh{} & \spo{}   & \pfyl{}  &\nce{} \\ \midrule
        \textbf{Deg $4$}   & $1.55 \pm 0.18$ & $611.47 \pm 23.52$ & $518.07 \pm 51.89$ & $196.96 \pm 35.92$ & $1220.68 \pm 85.39$ & $1328.99 \pm 28.87$ & $\bm{151.80} \pm 24.21$ \\
        \textbf{Deg $6$}   & $1.16 \pm 0.13$ & $502.71 \pm 16.03$ & $573.87 \pm 20.19$ & $253.93 \pm 27.67$ & $1191.29 \pm 42.63$ & $1456.21 \pm 34.18$ & $\bm{155.95} \pm 24.46$ \\
      \bottomrule
      \end{tabular}
      }
    \end{subtable}
    \label{tab:tsp50}
\end{table}

\subsubsection{CVRP.}
\begin{table}[ht]
  \centering
  \caption{Experimental Results for CVRP$20$}
    \begin{subtable}{\textwidth}
      \centering
      \caption{Average Test Normalized Regret (\%) with Standard Deviation}
      \setlength{\tabcolsep}{5pt}
      \resizebox{\columnwidth}{!}{
      \begin{tabular}{r|r|r|r|r|r|r|r}
      \toprule
        \textbf{Methods}     & 2-Stage       & \cavee{}      & \cavep{}      & \caveh{}      & \spo{}        & \pfyl{}       & \nce{} \\ \midrule
        \textbf{Deg $4$}     & $10.10 \pm 0.64$ & $9.26 \pm 1.56$ & $6.44 \pm 0.24$ & $7.92 \pm 0.52$ & $\bm{5.94} \pm 0.25$ & $6.32 \pm 0.28$ & $15.77 \pm 0.96$ \\
        \textbf{Deg $6$}     & $19.50 \pm 1.22$ & $11.64 \pm 0.25$ & $\bm{7.94} \pm 0.54$ & $11.44 \pm 1.14$ & $8.75 \pm 0.28$ & $8.09 \pm 0.57$ & $18.96 \pm 1.01$ \\
      \bottomrule
      \end{tabular}
      }
    \end{subtable} 

    \begin{subtable}{\textwidth}
      \centering
      \caption{Average Training Time (Sec) with Standard Deviation}
      \setlength{\tabcolsep}{7pt}
      \resizebox{\columnwidth}{!}{
      \begin{tabular}{r|r|r|r|r|r|r|r}
      \toprule
        \textbf{Methods}   & 2-Stage        & \cavee{}       & \cavep{}       & \caveh{}       & \spo{}         & \pfyl{}       & \nce{} \\ \midrule
        \textbf{Deg $4$}   & $1.65 \pm 0.48$ & $213.56 \pm 42.36$ & $153.56 \pm 11.08$ & $\bm{44.52} \pm 6.27$ & $7020.11 \pm 1043.05$ & $3773.31 \pm 288.84$ & $583.56 \pm 170.67$ \\
        \textbf{Deg $6$}   & $1.54 \pm 0.25$ & $208.95 \pm 12.90$ & $127.94 \pm 13.84$ & $\bm{51.83} \pm 8.78$ & $2204.83 \pm 99.86$ & $6197.84 \pm 288.63$ & $470.20 \pm 84.46$ \\
      \bottomrule
      \end{tabular}
      }
    \end{subtable}
    \label{tab:vrp20}
\end{table}

As mentioned earlier, we are the first to tackle a CVRP in an end-to-end predict-then-optimize setting. To solve CVRP as a BLP, we formulated the problem with k-path cuts~\cite{kohl19992} and solved it using Gurobi. For CVRP20,~\cavep{}, \spo{} and~\pfyl{} compete for the lowest regret, with~\caveh{} running closely behind. However,~\cavep{} completes its $10$ training epochs in roughly 2-3 minutes, whereas~\spo{} and~\pfyl{} require more than 1 hour each, on average. The regret of \nce{} is high and the training time has increased a lot. Similar as TSP, we also employ a linear relaxation of the BLP during training for \spo{} and \pfyl{} (see Appendix~\ref{subapdx:vrprel}), which results in a significant decrease in training time at the cost of regret.

\begin{table}[ht]
  \centering
  \caption{Experimental Results for CVRP$30$}
    \begin{subtable}{\textwidth}
      \centering
      \caption{Test Normalized Regret}
      \setlength{\tabcolsep}{5pt}
      \resizebox{0.6\columnwidth}{!}{
      \begin{tabular}{r|r|r|r|r|r|r|r}
      \toprule
        \textbf{Methods}   & 2-Stage  & \cavee{} & \cavep{} & \caveh{} & \spo{}   & \pfyl{} & \nce{} \\ \midrule
        \textbf{Deg $4$}   & 0.1972   & 0.1254   & \textbf{0.0913}  & 0.0999   & \multicolumn{2}{|c|}{N/A} & 0.1828 \\
      \bottomrule
      \end{tabular}
      }
    \end{subtable} 

    \begin{subtable}{\textwidth}
      \centering
      \caption{Training Time (Sec)}
      \setlength{\tabcolsep}{7pt}
      \resizebox{0.6\columnwidth}{!}{
      \begin{tabular}{r|r|r|r|r|r|r|r}
      \toprule
        \textbf{Methods}   & 2-Stage  & \cavee{} & \cavep{} & \caveh{} & \spo{}   & \pfyl{} & \nce{} \\ \midrule
        \textbf{Deg $4$}   & 9.27     & 331.73    & 287.77  & \textbf{132.62}   & \multicolumn{2}{|c|}{$\geq 100$h} & 884.95 \\
      \bottomrule
      \end{tabular}
      }
    \end{subtable}
    \label{tab:vrp30}
\end{table}

This performance gap is even more pronounced for CVRP30 in Table~\ref{tab:vrp30}. It takes roughly 20 seconds on average to solve a single CVRP30 instance, thus requiring 8 or so hours to traverse the entire dataset of 1,000 training instances once for \spo{} and \pfyl{}. This makes end-to-end training with these methods impractical for real-world applications. In contrast,  \cave{} demonstrates its ability to handle such a challenging problem efficiently. Note that due to the scale of the problem, our experimental evaluation was not repeated with random seeds and we used a smaller test set comprising only $10$ instances. All~\cave{} variants achieve test regrets of $9$-$12\%$ compared to the 2-Stage method's far higher $20\%$, while requiring only 2-6 minutes of training each. In addition, \nce{} achieves $18\%$ with $15$ minutes of training and is thus worse than all variants ~\cave{} in both metrics. To our knowledge, this is the hardest optimization problem ever targeted in the predict-then-optimize literature, a feat that is only possible due to the computational efficiency brought about by~\cave{}.

\section{Conclusion} \label{sec:conclu}

\cave{} reframes the end-to-end training problem for predict-then-optimize as a regression task. Unlike the traditional two-stage approach, which regresses on the cost vectors, our framework instead regresses on cones that correspond to optimal solutions under the true costs. \cave{} can be seen as an attempt at obtaining the best of both worlds: fast training with a regression loss that does not require solving hard integer optimization problems in every iteration of gradient descent, and a loss function that penalizes cost predictions that point in the wrong direction relative to the optimal decision. We proposed three versions of our method with varying performance trade-offs. 

The best of the three appears to be~\cavep{}, which regresses on an inner vector of the optimal subcone of a training instance, resulting in stable and efficient training, as well as test \text{regret} results that compare to state-of-the-art methods that require training 30 times longer on CVRP20, or do not even complete a single training epoch in 8 hours for CVRP30. We note that if we were to use early termination during training,~\cave{} methods would record even smaller training times as they do converge in fewer gradient descent iterations than competing methods as per Fig.~\ref{fig:loss}. We hope that our framework will enable the adoption of end-to-end predict-then-optimize in a wider range of applications and have made our implementation available with plans to make~\cave{} one of the standard methods within the PyEPO package following the publication of this work.

\cave{} is limited to binary problems. In practice, this is not too problematic, as bounded integer variables can be represented using a set of binary variables. Another limitation of our method is the lack of theoretical guarantees. In particular, we currently do not know whether the loss function in Eq.~\eqref{eq:loss} or a modification thereof could be proven to be a valid upper bound on \text{regret}, as does the \spo{} loss of~\citet{elmachtoub2021smart} for example. This direction merits further investigation. An interesting connection may be established between~\cave{} and recent ML methods such as~\cite{misra2022learning} to predict the active constraints of a family of similar optimization problems for which optima are known. Rather than predict-then-optimize, the goal of~\citet{misra2022learning} is to accurately predict the active set to solve a reduced optimization problem over only that set.

%
%
%
\bibliographystyle{splncs04nat}
\bibliography{ref}
\appendix

\section{More Experiments} \label{apdx:exper}

\subsection{TSP Relaxation.} \label{subapdx:tsprel}

To enhance training efficiency, it has been proposed in \cite{mandi2020smart} that a linear relaxation can be used as a substitute for the original BLP during training. Because the DFJ formulation utilizes constraint generation to handle subtour elimiation constraints, it is challenging to achieve linear relaxation due to the exponential number of constraints. In this study, we used the LP relaxation of the Miller-Tucker-Zemlin (MTZ) formulation \cite{miller1960integer} of the TSP, and trained \spor{} and \pfylr{} on the same TSP instances with 20 and 50 nodes. Although relaxation methods are more efficient than \cave{} in TSP20, they result in higher regret. As the size of the model increases, the relaxation approach also loses its efficiency advantage.

\begin{table}[htbp]
  \centering
  \caption{Experimental Results for TSP$20$ Relaxation}
  \begin{minipage}{0.48\textwidth}
    \centering
    \subcaption{Average Test Normalized Regret (\%) with Standard Deviation}
    \setlength{\tabcolsep}{6pt}
    \begin{tabular}{r|r|r}
      \toprule
      \textbf{Methods} & \spor{}  & \pfylr{} \\\midrule
       \textbf{Deg $4$} & $7.75 \pm 0.32$ & $9.15 \pm 0.50$ \\
       \textbf{Deg $6$} & $9.83 \pm 0.57$ & $11.28 \pm 0.90$ \\
      \bottomrule
    \end{tabular}
  \end{minipage}
  \hfill
  \begin{minipage}{0.48\textwidth}
    \centering
    \subcaption{Average Training Time (Sec) with Standard Deviation}
    \setlength{\tabcolsep}{6pt}
    \begin{tabular}{r|r|r}
      \toprule
      \textbf{Methods} & \spor{}  & \pfylr{} \\\midrule
       \textbf{Deg $4$} & $71.92 \pm 2.05$ & $77.25 \pm 0.60$ \\
       \textbf{Deg $6$} & $67.99 \pm 0.60$ & $53.32 \pm 3.74$ \\
      \bottomrule
    \end{tabular}
  \end{minipage}
  \label{tab:tsp30rel}
\end{table}

\begin{table}[htbp]
  \centering
  \caption{Experimental Results for TSP$50$ Relaxation}
  \begin{minipage}{0.48\textwidth}
    \centering
    \subcaption{Average Test Normalized Regret (\%) with Standard Deviation}
    \setlength{\tabcolsep}{6pt}
    \begin{tabular}{r|r|r}
      \toprule
      \textbf{Methods} & \spor{}  & \pfylr{} \\\midrule
       \textbf{Deg $4$} & $10.17 \pm 0.23$ & $11.11 \pm 0.33$ \\
       \textbf{Deg $6$} & $13.14 \pm 0.46$ & $13.38 \pm 0.58$ \\
      \bottomrule
    \end{tabular}
  \end{minipage}
  \hfill
  \begin{minipage}{0.48\textwidth}
    \centering
    \subcaption{Average Training Time (Sec) with Standard Deviation}
    \setlength{\tabcolsep}{6pt}
    \begin{tabular}{r|r|r}
      \toprule
      \textbf{Methods} & \spor{}  & \pfylr{} \\\midrule
       \textbf{Deg $4$} & $386.06 \pm 9.69$ & $536.67 \pm 4.94$ \\
       \textbf{Deg $6$} & $636.99 \pm 3.04$ & $510.37 \pm 3.46$ \\
      \bottomrule
    \end{tabular}
  \end{minipage}
  \label{tab:tsp50rel}
\end{table}

\subsection{CVRP Relaxation.} \label{subapdx:vrprel}

Similarly, the CVRP formulation with k-path cuts also struggles to obtain a linear relaxation. Thus, the MTZ formulation can be modified with constraints
$$
u_j - u_i \geq Q (x_{ij} - 1) + q_j \quad \forall i \neq j, i \neq 0, j \neq 0
$$
for subtour elimilation and customer demand, where $Q$ is the capacity and $q_i$ is the demand of customer $i$. The \spor{} and \pfylr{} are trained for the same VRP instances with 20 customers. While the linear relaxation approach demonstrates high efficiency in these scenarios, it does not guarantee a low regret compared to \cave{}.

\begin{table}[htbp]
  \centering
  \caption{Experimental Results for CVRP$20$ Relaxation}
  \begin{minipage}{0.48\textwidth}
    \centering
    \subcaption{Average Test Normalized Regret (\%) with Standard Deviation}
    \setlength{\tabcolsep}{6pt}
    \begin{tabular}{r|r|r}
      \toprule
      \textbf{Methods} & \spor{}  & \pfylr{} \\\midrule
       \textbf{Deg $4$} & $8.03 \pm 0.38$ & $17.07 \pm 0.63$ \\
       \textbf{Deg $6$} & $15.73 \pm 0.39$ & $19.19 \pm 1.66$ \\
      \bottomrule
    \end{tabular}
  \end{minipage}
  \hfill
  \begin{minipage}{0.48\textwidth}
    \centering
    \subcaption{Average Training Time (Sec) with Standard Deviation}
    \setlength{\tabcolsep}{6pt}
    \begin{tabular}{r|r|r}
      \toprule
      \textbf{Methods} & \spor{}  & \pfylr{} \\\midrule
       \textbf{Deg $4$} & $78.95 \pm 0.73$ & $78.80 \pm 1.19$ \\
       \textbf{Deg $6$} & $78.74 \pm 3.82$ & $81.80 \pm 0.86$ \\
      \bottomrule
    \end{tabular}
  \end{minipage}
  \label{tab:vrp20rel}
\end{table}

\end{document}